# Enhancing Scientific Reproducibility Through Automated BioCompute Object Creation Using Retrieval-Augmented Generation from Publications

Sean Kim[1*] and Raja Mazumder[1,2^]

[1]The Department of Biochemistry & Molecular Medicine, GW School of Medicine and Health Sciences, Washington DC 20037, United States of America
[2]The McCormick Genomic and Proteomic Center, The George Washington University, Washington, DC 20037, United States of America

*Corresponding author
^Co-corresponding author

Sean Kim: skim658@gwu.edu
Raja Mazumder: mazumder@gwu.edu

Abstract

The exponential growth in computational power and accessibility has transformed the complexity and scale of bioinformatics research, necessitating standardized documentation for transparency, reproducibility, and regulatory compliance. The IEEE BioCompute Object (BCO) standard addresses this need but faces adoption challenges due to the overhead of creating compliant documentation, especially for legacy research. This paper presents a novel approach to automate the creation of BCOs from scientific papers using Retrieval-Augmented Generation (RAG) and Large Language Models (LLMs). We describe the development of the BCO assistant tool that leverages RAG to extract relevant information from source papers and associated code repositories, addressing key challenges such as LLM hallucination and long-context understanding. The implementation incorporates optimized retrieval processes, including a two-pass retrieval with re-ranking, and employs carefully engineered prompts for each BCO domain. We discuss the tool's architecture, extensibility, and evaluation methods, including automated and manual assessment approaches. The BCO assistant demonstrates the potential to significantly reduce the time and effort required for retroactive documentation of bioinformatics research while maintaining compliance with the standard. This approach opens avenues for AI-assisted scientific documentation and knowledge extraction from publications thereby enhancing scientific reproducibility. The BCO assistant tool and documentation is available at https://biocompute-objects.github.io/bco-rag/.

Background

Fueled by the exponential growth in compute power and resource accessibility in the 21st century, the field of bioinformatics continues to move at an unprecedented pace of discovery and experimentation[1, 2]. This is clearly evidenced by the substantial amount of recent research done in leveraging modern generative artificial intelligence (AI) for future scientific advancements and innovation. However, there has been less research aimed at preserving the transparency and portability of existing research. As bioinformatics workflows become increasingly complex, they will only retain their value to be built and iterated upon if proper documentation and source material exist[3].

The BioCompute Object (BCO)[4, 5] project is an Institute of Electrical and Electronics Engineers (IEEE) community-driven open standards framework for standardizing and sharing computations and analyses. The BCO standard's development was initially driven by the rapid increase in next-generation sequencing data and the complexity of workflows used to analyze and transform this data[6-8]. Since then, it has been shown that BCOs can be used to document other bioinformatics workflows[9-11]. As bioinformatics processes become more intricate and complex, documentation standardization is imperative not only for general experimental preservation, transparency, accuracy, automation, and reproducibility but also for communications between researchers at academia, regulatory agencies, and pharmaceutical companies.

The BioCompute Object standard defines documentation as JavaScript Object Notation (JSON). For a given project BCO, information critical to reproducing and understanding the workflow is organized into the following domains: the Provenance domain, the Usability domain, the Extension domain, the Description domain, the Execution domain, the Parametric domain, the Input and Output domain, and the Error domain[12]. Each domain is meant to logically organize information for a reviewer to easily gain a clear understanding of what the workflow does, how it works, and how to reproduce it.

As with any standard, and especially so with a documentation standard, widespread adoption is largely hindered by two main factors: the overhead required to maintain standard compliance and the work required to retroactively meet compliance with past and legacy research. To reduce the complexity of assuring compliance with the BCO standard, the BioCompute Object Portal[13] was developed. The portal provides a graphical user interface, automatic JSON schema validation, and granular domain guidance. Although much more user-friendly than manually curating and validating a complex JSON object, the portal still requires significant work from the end user to collect, extract, and identify the relevant information for each domain.

There are two main scenarios when attempting to document research in compliance with the BCO standard. The first scenario applies to documenting research that is in progress. In this case, it is much more likely the details of the research workflows are readily available, making BCO compliance relatively straightforward using the BCO Builder[14] or the academic[15] and commercial tools cited on the website[13, 16]. These tools can be used to create comprehensive BCOs that can serve as the project's main documentation. This approach reduces the potential duplication of documentation effort as BCOs can also be easily converted to more human-readable formats such as markdown[17]. The second case applies to retroactively documenting research that has already been completed. In this case, multiple factors can greatly influence the overhead required to create comprehensive BCO documentation. These factors include time elapsed since the research was completed, whether primary researchers are still available, and the level of detail in the existing documentation. In many instances, the only readily available source of information will be the published scientific paper and the legacy code repository, making BCO compliance much more difficult, labor-intensive, and time-consuming.

The primary objective of this project is to demonstrate the viability of an automated BCO creation pipeline from scientific papers to reduce the overhead in retroactively documenting pre-existing research in compliance with the BCO standard. With the recent advancements in transformer-based large language models (LLMs), the feasibility and utility of an automated BCO creation assistant presents a practical and intriguing use case.

Implementation

*Considerations*
In preparation for designing the architectural foundation for this proof of concept, several fundamental questions about the viability of this use case in the context of LLMs arise. With further research on the capabilities, strengths, and weaknesses of the current frontier LLMs, it becomes clear this use case is somewhat contradictory to what general-purpose LLMs were originally designed for.

Traditional natural language processing models were trained to be precise experts, with a focus on a particular domain, as evidenced by the narrow scope of their original training datasets. As the first modern, transformer-based LLMs were introduced, the breakthrough milestone was characterized by OpenAI as a "move towards more general systems which can perform many tasks"[18]. This shift in expectation, from models that perform highly specialized tasks, to models that perform well at generalizing a large array of potential tasks, demonstrates the fundamental hurdle in designing this proof-of-concept tool. Modern frontier LLMs were designed for creative, flexible, free-text responses that represent plausible natural language in any context. BCOs, on the other hand, were designed for deterministic, non-ambiguous, structured, and fundamentally reproducible documentation, not creative output. This presents a significant challenge where the LLM's inherent ability to extrapolate on generic circumstances has to be steered in favor of strict adherence to the source material.

This leads to our next consideration. While the most powerful general-purpose LLMs have an impressive ability to generalize and extrapolate information outside of their original training context, it is not always to the user's benefit. A well-known but difficult-to-solve problem with general-purpose LLMs is their tendency to hallucinate[19, 20]. Although significant advancements have been made in modern LLM capabilities, the driver of these advancements has been scale, defined by the number of model parameters, dataset size, and amount of compute[21]. Given this push towards massive scale over fundamental architectural and algorithmic changes in model design, even the largest state-of-the-art LLMs can only mitigate hallucination. The current most powerful open source LLM, Llama 3.1 405B Instruct, was deliberately designed with only minor adaptations on the standard dense transformer architecture to maximize the ability to scale[22]. The training process was completed at a scale previously unprecedented for an open-source model, using 16,000 H100 GPUs and approximately 30 million GPU hours to complete training. Despite the relative simplicity in architecture, the complexity at scale makes gaining insight into these models a significant task[23].

The last question in addressing the viability of our implementation strategy is the practicality of working with arbitrarily long source papers as inputs. Research has shown that standalone LLMs can struggle with long contexts, particularly when they must access relevant information in the middle of long contexts. Performance was found to degrade when changing the position of relevant information within the LLMs context window, indicating current LLMs do not comprehensively make use of information in long input contexts[24, 25]. This presents a clear challenge to creating BCOs from arbitrarily long scientific papers. For example, the Usability domain represents a high-level plain language description of what was done in the workflow, analogous to an abstract. The Description domain represents the workflow at a lower level, documenting the exact pipeline steps with software prerequisites and any input or output files for each step. As a result, it is likely the relevant information for the Usability domain will be present at the beginning of the source paper whereas Description domain-related information will be closer to the middle of the paper. Given this, it is reasonable to expect higher quality output for the Usability domain than the Description domain when ingesting the entire paper as part of the context window in a standalone LLM approach (given the paper fits within the model's maximum input window).

*Architecture*
In searching for solutions to the main use case considerations, Retrieval-Augmented Generation (RAG) was identified as a method that can address many of our initial challenges. As discussed previously, pre-trained standalone LLMs that rely on parameterized implicit knowledge have inherent limitations. They cannot easily expand or revise memory, it is difficult to gain insight into their predictions, and they can more easily produce hallucinations. Retrieval-Augmented Generation aims to address these issues by proposing a hybrid model that combines parametric memory with dynamic access to non-parametric knowledge. This inclusion of non-parametric, or external knowledge, addresses our considerations on multiple fronts. RAG systems can have their knowledge directly revised and expanded, and knowledge can be inspected and interpreted[26]. The RAG approach also helps to narrow the scope of the LLM, helping to prevent false extrapolation and potential hallucinations.

The high-level initial architecture for the BCO assistant is described in Figure 1. The workflow starts from the raw data source, which in our case is the scientific paper to generate a BCO. From the raw data source, a data loader is used to extract the information from the source paper, depending on the loader technique this can include extracting text, figures, images, and other metadata. The resulting extracted data is then partitioned into chunks, also known as nodes. The resulting chunks are embedded, a process where the text is converted into a vector representation, which encodes the text's semantic information in high-dimensional latent space. The embeddings are then stored in a vector database. The stored embeddings form the core of the RAG system, allowing for semantic retrieval from the vector store based on the similarity to the user prompt. When a prompt is fed into the pipeline, the prompt also goes through the embedding process. The cosine similarity between the query embedding and each of the nodes is calculated using the dot product of the normalized node embeddings. The similarity scores for each node are sorted in descending order and the top-k most similar nodes are returned. These nodes are then passed to the LLM along with the original prompt to supplement the LLM with relevant contextual information.

In using the BCO assistant, users must generate each BCO domain separately. The reasons for this constraint are multi-faceted. BCO domains are already logically organized in self-containing conceptually related sections[16]. Given the domain organization within a complete BCO, the relevant information to create each domain would be contained in different sections of the source paper. This inherent logical structure lends itself naturally to the retrieval concept within RAG systems. Assuming a capable LLM is being used, the retrieval step is the next most important process for high-quality responses. Separating the generation of each domain leads to the most accurate information retrieval. The second primary reason is, as we simultaneously gathered data on user feedback and evaluation, a per-domain approach was found to be more engaging for the user. The user is more likely to comprehensively verify more concise per-domain outputs instead of interpreting a potentially large and complex one-shot JSON response. Lastly, the requirements for each domain vary significantly. What makes for a strong Usability domain does not meet the same semantic requirements as a comprehensive Parametric domain. Gathering more granular evaluation data allows for deeper insights into the strengths and weaknesses of the tool.

At the core of the BCO assistant is the LlamaIndex Python library[27]. The initial implementation of the tool used many high-level components from the LlamaIndex library, including the Document container, custom Node parsers, LlamaIndex community data Connectors, and the in-memory base Vector Store Index component.

*Improvements*

After the initial tool implementation showed promising potential, the next step was to optimize the different components of the RAG pipeline. As a starting point, we investigated the lowest cost optimization in any LLM workflow, prompt engineering. Although RAG itself can be seen as a prompt engineering method[28], there are additional considerations in how to format RAG prompts as we must consider not only the eventual LLM parsing but the retrieval process. Given the narrow focus of the tool, it was reasonable to restrict free text prompting by the user in favor of standardized and comprehensive prompts specific to each domain. The initial prompt's importance in the quality of the retrieval process is critical for the quality of the generated domain. A simple user prompt in the form of "generate the usability domain" will lack the semantic information required for the retrieval process to accurately identify relevant data to provide to the LLM. We started by testing meticulously curated prompts that follow a standardized structure: the request statement (request for the particular BCO domain), domain description (a plain language description of what the domain represents and what type of information it entails), likely sections of the paper the information will be in, and the JSON schema for the output structure.

As prompts were continually tested, it became clear that there was an inherent tradeoff between optimizing for the retrieval process versus the LLM request. In optimizing the prompt for the retrieval process, the resulting LLM request is polluted with unnecessary information regarding the semantics of the document retrieval. On the other hand, including the domain's JSON schema for the output formatting polluted the retrieval process in calculating and evaluating similarity scores. Splitting up the retrieval embedding and the LLM prompt improved response consistency, reproducibility, accuracy to conforming with domain schemas, and overall content quality. The standardized base Usability domain LLM prompt is shown in Figure 2a. Figure 2b shows the Usability domain retrieval prompt. Figure 2c provides an example of a generated Usability domain from a publication[29].

Further focus was directed at optimizing the accuracy of the retrieval process. The existing similarity search used the pre-computed embeddings, making it fast and inexpensive in a first-pass retrieval for a large top-k value[30]. Traditional bi-encoder models are highly efficient at capturing semantic information through embeddings. However, they encode the query and vector store documents independently. This independent encoding process can lead to contextual information loss as it does not allow the model to assess how specific elements of the query align with corresponding elements in the documents. To address this, a re-ranking step was added after the initial retrieval process, as shown in Figure 3. Research has shown the advantages of first efficiently retrieving a larger amount of relevant data followed by a more computationally intensive re-ranking step using a cross-encoder model[31, 32]. Unlike the initial retrieval process, the cross-encoder re-rank model processes the prompt and retrieved documents directly, rather than the pre-computed embeddings. The cross-encoder model evaluates the query and each retrieved node together in a single forward pass, allowing for token-level interactions between the prompt and the retrieved text. By only using this more computationally expensive step on a smaller corpus of data, we retain performance while simultaneously improving output quality.

Once the retrieval process was optimized, we explored inherent limitations in source papers as the sole information source for full workflow documentation. While most published papers include the high-level information required to document the experimental process, there were still limitations that appeared frequently throughout testing. A common scenario is for authors to include links to a GitHub repository that contains the source code and corresponding input and output files for the workflow. This linkage to external documents limits the tool's ability to create comprehensive BCOs from the source paper alone. For example, the Description domain should thoroughly document each pipeline step for the workflow including any dependencies and locations of any corresponding input or output files. The Parametric domain should include the exact run parameters for the corresponding steps in the Description domain. This level of granular detail is often documented in the repository files and not explicitly described in the published paper. We observed that fields that require knowledge from external dependencies will result in non-specific information. To mitigate this common scenario, an optional preprocessing step was built to allow users to include a link to a GitHub repository as part of the data load step. Provided the repository is not private, the contents can be automatically fetched, chunked, and embedded for the vector store along with the source paper, expanding the retrieval process's knowledge pool.

*Extensibility*
Given the proof-of-concept nature of the project, it was important to retain maximum modularity and extensibility as different parameter sets for the pipeline are continuously being tested and evaluated. Additional configuration options and parameter values can be easily added and removed as further testing is completed. When running the tool in its standard configuration, the user will first be prompted to choose the various parameter configurations for the RAG pipeline. This includes the type of data loader, the chunking strategy, the embedding model, the similarity top k value, and the LLM model.

The project also includes high-level wrappers for automated testing at scale. Inspired by hyper-parameter optimization techniques in traditional machine learning[33], random and grid search tools were built to allow for extensive automated testing for different parameter configurations. Users can define a search space, and the parameter search wrapper will automatically abstract the building and configuring of different RAG pipelines for different trials. The resulting trials are completed sequentially and produce each domain for its corresponding parameter set.

While the parameter search tool allows for simple automated testing of many different configuration sets, the evaluation of these results is a challenge. An automated tester was built using DeepEval[34], an open-source LLM evaluation framework. DeepEval offers a variety of pre-configured metrics for automated evaluation of RAG pipelines. The BCO assistant's automated test suite tests for answer relevancy and faithfulness. The answer relevancy metric uses the prompt and the generated domain to evaluate how relevant the final generated output is to the original input prompt. The metric attempts to quantify relevancy, does the generated content directly relate to the question at hand, appropriateness, is the content appropriate given the context of the input, and focus, does the content stays on topic. In evaluating answer relevancy, DeepEval first uses an LLM to extract the statements made in the generated output and then uses the same LLM to classify whether each statement is relevant to the input prompt[34]. The faithfulness metric uses the prompt, generated domain, and the retrieved nodes to assess how accurate and truthful the finalized generated output is concerning the source material. The metric attempts to ensure that the content is relevant, factual, and does not contradict the information gathered from the retrieval step. In evaluating faithfulness, DeepEval first uses an LLM to extract all the claims made in the generated output and then uses the same LLM to classify whether each claim is truthful based on the facts presented in the retrieved nodes[34]. Although automated evaluation is useful in preliminary testing, each BCO domain is inherently nuanced in the content and format requirements making reliable and consistent automated evaluation difficult.

To assist with this, a simple evaluation application was built to ease comprehensive evaluation at scale and to standardize the metrics captured for each generated domain. The evaluation tool is shown in Figure 4. The main screen shows the evaluator the human-curated domain along with the RAG-generated domain. From there, the "Source Nodes" tab can be used to view the retrieved source nodes that were passed as context to the LLM along with the domain prompt. The "Parameter Set" tab displays the exact parameter configuration for the run including the GitHub repository, the repository branch that was indexed, and any file extension or directory filters imposed during the data ingestion. The "Evaluate" tab contains a logical grouping of different scoring categories (Figure 5).

Discussion

*Tool Overview*
The BCO assistant empowers researchers to continue working on pushing the boundaries of science by automating the time-consuming process of documenting legacy research. The assistant also allows users unfamiliar with legacy research to generate IEEE standards-compliant documentation at scale, helping them to better understand and compare legacy research during their experimental processes. Figure 6 illustrates the user workflow steps to generate a BCO domain. The tool is started through a command line interface. Once started, the user will be asked to choose their parameter configurations. Once the configuration options are selected, the user will have the opportunity to specify a GitHub repository to include as supplementary data for the indexing process. If a GitHub repository is included, additional options will be presented to the user to select which repository branch to index, and any directory or file extension filters to apply. Each filter can be specified as an inclusion or exclusion filter. Once the resulting RAG system is built, the user will be prompted for which BCO domain they would like to generate.

*Step-by-step instructions*
The minimum prerequisites to using the BCO assistant are a PDF of a paper, an OpenAI API key, and at least Python version 3.10. At a high level, prospective users can clone the code repository onto their local machines, add their target PDF file to the papers directory, add their OpenAI API key to an environment file, and start the program. Once started, users will be prompted to select any custom configuration options or can simply select the default options. At each step of the configuration menus, links are provided to detailed documentation on the differences, strengths, and weaknesses of the different configuration options. The BCO assistant supports any operating system that can run Python and can be run on any basic command line interface. Step-by-step instructions on how to use the tool are provided at [https://biocompute-objects.github.io/bco-rag/installation/](https://biocompute-objects.github.io/bco-rag/installation/).

*Future Directions*
The core components in the future could be split into a microservices architecture ready for deployment at scale. A microservices architecture will not only allow for future scaling but also increase the ability to experiment. At this phase, experimentation methods such as A/B testing, bandit testing, user surveys, and feature flags will help us to identify the highest-performing versions of the BCO assistant. Once we have a better understanding of the tool's strengths and weaknesses in real-world testing, the system can be further fine-tuned using open source LLMs.

An additional avenue of improvement is the recent research done on the impact of format restrictions on the quality of LLM responses[35]. The research explores how enforcing strict structured output constraints (such as JSON) can lower the quality of the output content by hindering the LLM's reasoning abilities. This research is particularly impactful for the BCO assistant as we are not only imposing JSON outputs but also constraining the LLM to produce JSON that is valid against a complex JSON schema. Similar to the multi-phase process of RAG, exploration could be done in implementing a two-phase LLM step where the first LLM pass is only responsible for aggregating the required information, and the second LLM (whether a different model or a call to the same model) is only used for output formatting.

Although the primary use case for the BCO assistant is retroactively documenting pre-existing research, the BCO assistant also provides functionality for documenting in-progress research. The BCO standard is comprehensive, requiring high minimum requirements for conformance. Due to the comprehensive nature of BCOs, in-progress documentation can be generated with the loosened format and content requirements, generating a markdown document with sections covering similar content to the BCO domains. Once the research is completed, the BCO assistant will recognize the in-progress documentation and prompt the user to include this documentation in the ingestion process, further improving the relevant documentation available during the final retrieval process.

*Limitations*

This project has some limitations. First, due to cost and resource constraints, we were unable to test the latest frontier open-source models such as the Llama 3.1 70 or 405B models. Due to these constraints, we were limited to leveraging compute offered by companies such as OpenAI through their public APIs. The second limitation is the dependency on external libraries. External libraries such as LlamaIndex were leveraged to abstract some of the low-level implementations of the RAG framework. While libraries such as LlamaIndex are powerful tools in general RAG pipelines, further evaluation of the tool could warrant a specialized approach where custom components should be built. However, with the extensible design of the BCO assistant, further improvements and other LLMs can be tested and swapped out with limited development overhead by other users or the original developers.

Conclusion

As compute resources become more accessible and powerful, the field of bioinformatics continues to push the limits of life and computer sciences. Standards like BioCompute are essential to the continued sustainability of the collective progress of bioinformatics methods. The BCO assistant project demonstrates a novel application of retrieval-augmented generation to address the challenge of retroactively documenting bioinformatics research in compliance with the BCO standard. Our implementation highlights the potential of AI-assisted documentation while the modular architecture and focus on extensibility allows for continued refinement and optimization of the tool. As the tool matures, it has the potential to play a crucial role in facilitating knowledge transfer and supporting the ongoing advancements of bioinformatics as a field.

Code Availability

All the source code for this project is available on the GitHub repository (https://github.com/biocompute-objects/bco-rag). The full project documentation, including source code documentation, can be found on the documentation page (https://biocompute-objects.github.io/bco-rag/).

Acknowledgments
We would like to thank Jonathon Keeney, Patrick McNeely, Chinweoke Okonkwo, Tianyi Wang, and Charles Hadley King for providing valuable comments.

**Figure Legends**

Figure 1. BCO assistant pipeline workflow. Illustration of the general architecture and workflow for the BCO assistant.

Figure 2. Standardized Usability domain prompts and output. 2a) The standardized Usability domain base prompt for used for the LLM prompting. 2b) The standardized Usability domain prompt for used for the retrieval process. 2c) This figure shows a generated Usability domain using the standardized prompts from the paper "High resolution measurement of DUF1220 domain copy number from whole genome sequence data" (https://pubmed.ncbi.nlm.nih.gov/28807002/).

Figure 3. BCO-RAG Pipeline workflow with retrieval re-ranking. This figure illustrates the workflow for the BCO-RAG tool with two-pass retrieval re-ranking.

Figure 4. BCO-RAG evaluation tool home screen. This figure shows the home screen of the evaluation application.

Figure 5. BCO-RAG evaluation tool evaluation screen. This figure shows the evaluation screen of the evaluation application.

Figure 6. BCO-RAG user workflow. This figure illustrates the workflow for the BCO-RAG tool from the user perspective.

Figure 1.

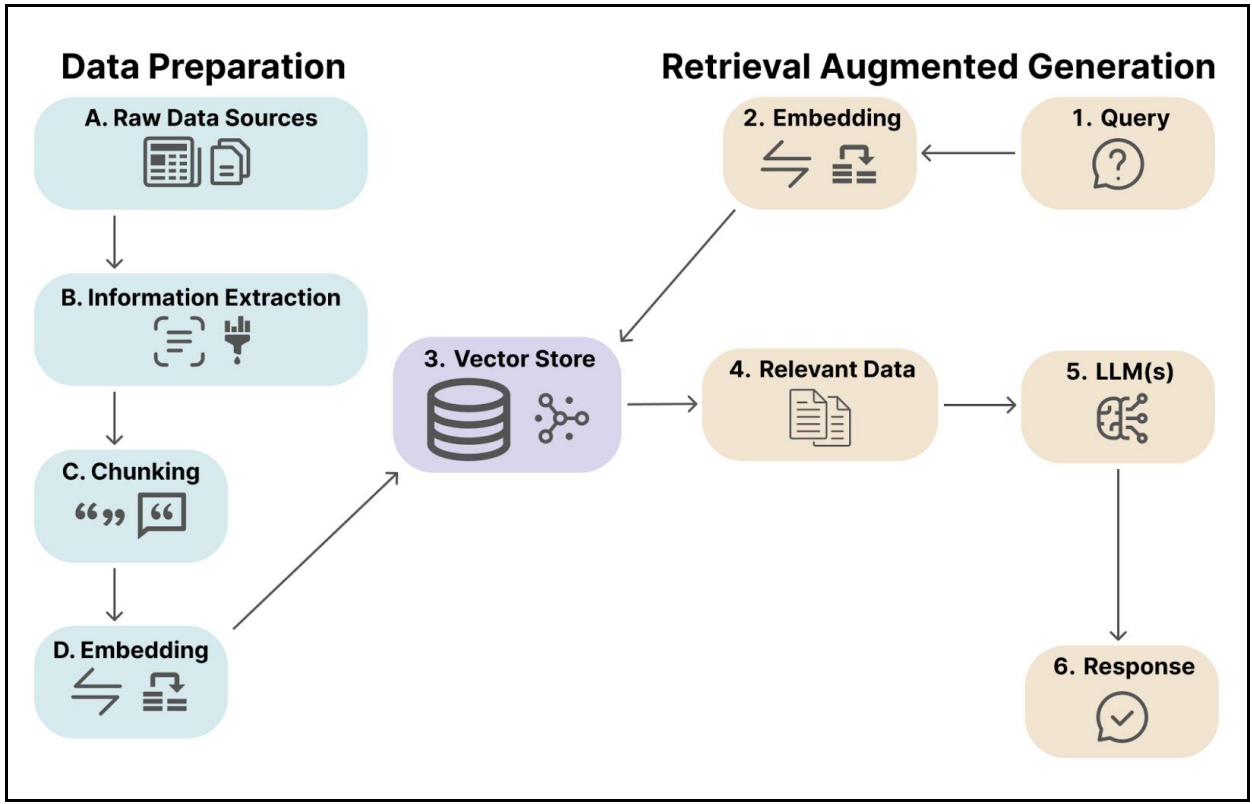

Figure 2a.

Can you give me a BioCompute Object (BCO) usability domain using the provided information from a bioinformatics workflow's documentation. The return response must be valid JSON and must validate against the JSON schema I am providing you. If the information for a field is not provided, leave it blank, do not make up any information. Do not repeat the JSON schema in your response, just make sure your response conforms against the schema. Please check your work before finalizing your response. The style of writing for the free text response fields should be similar to the style of a scientific paper, it should be written in past tense and not use any first person references. The usability domain is a high level, plain language description of the purpose and overall goal of the project workflow. It is analogous to an abstract. Keep the information high level and representative of the provided project paper and documentation excerpts.

The JSON schema is as follows:
```
{
   "$schema": "http://json-schema.org/draft-07/schema#",
   "$id": "https://w3id.org/ieee/ieee-2791-schema/usability_domain.json",
   "type": "array",
   "title": "Usability Domain",
   "description": "Author-defined usability domain of the IEEE-2791 Object. This field is to aid in search-ability and provide a specific description of the function of the object.",
   "items": {
      "type": "string",
      "description": "Free text values that can be used to provide scientific reasoning and purpose for the experiment",
      "examples": [
         "Identify baseline single nucleotide polymorphisms SNPs [SO:0000694], insertions [so:SO:0000667], and deletions [so:SO:0000045] that correlate with reduced ledipasvir [pubchem.compound:67505836] antiviral drug efficacy in Hepatitis C virus subtype 1 [taxonomy:31646]",
         "Identify treatment emergent amino acid substitutions [so:SO:0000048] that correlate with antiviral drug treatment failure",
         "Determine whether the treatment emergent amino acid substitutions [so:SO:0000048] identified correlate with treatment failure involving other drugs against the same virus"
      ]
   }
}
```

Figure 2b.

> Please retrieve the information required to produce a usability domain for a BioCompute Object (BCO). The usability domain in a BCO is a plain language description of what was done in the project or paper workflow. The information should convey the purpose of the paper and is a high level representation of the project workflow towards its purpose, not the low level details of the research. The usability domain information is likely to be in paper sections such as the abstract, background, summary, conclusions, or any section that describes the purpose and background of the project.

Figure 2c.

[
"This study aimed to explore the accuracy and precision of copy number estimation for DUF1220 domains within NBPF genes using high-resolution sequencing data. The methodology involved aligning longer reads with increased specificity to quantify individual domains and DUF1220 sequences. The approach was validated through simulations and digital droplet PCR (ddPCR), and applied to data from the 1000 Genomes Project. The results demonstrated the ability to accurately estimate DUF1220-clade specific copies and delineate clades within individual NBPF genes, providing insights into gene and intragenic domain copy number variations. This method enhances the capability to analyze the role of DUF1220 sequences in human variation and disease, and can be applied to other multi-copy gene families."
]

Figure 3.

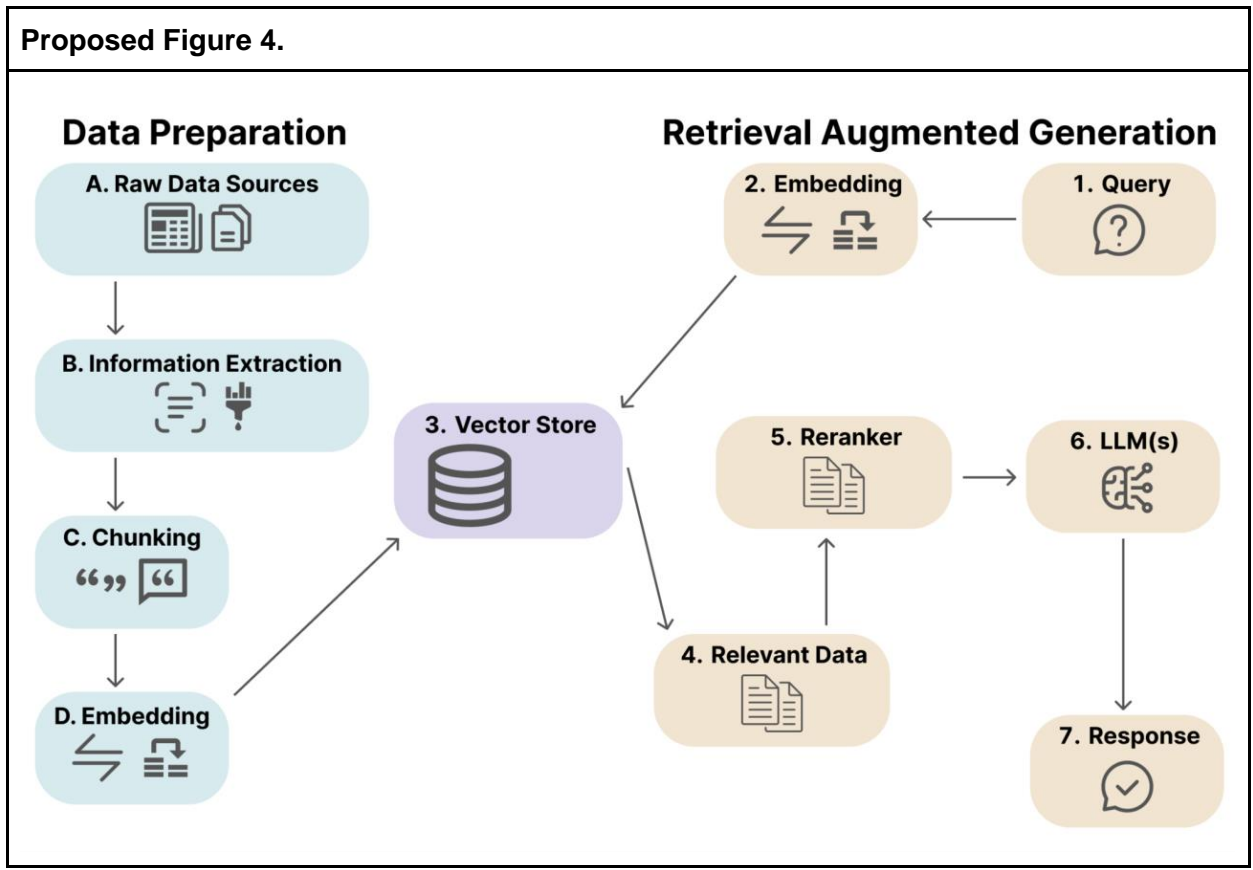

Figure 4.

Figure 5.

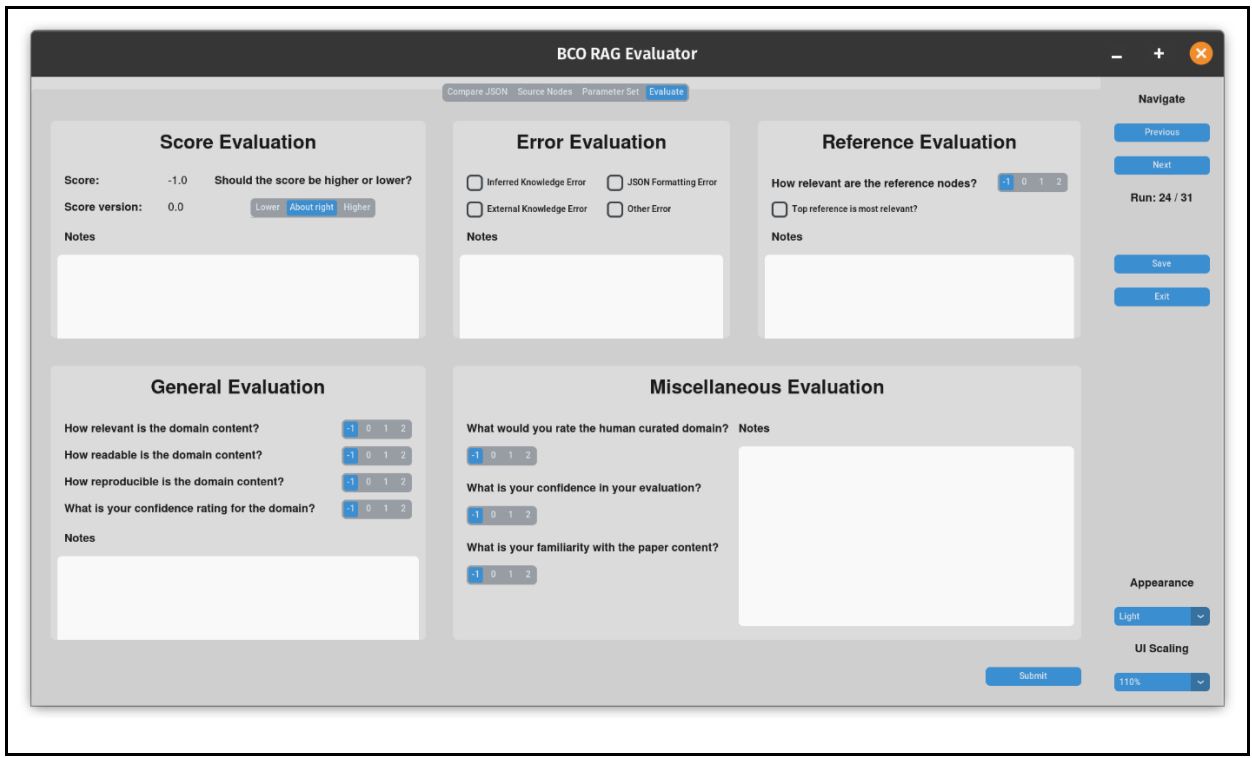

Figure 6.

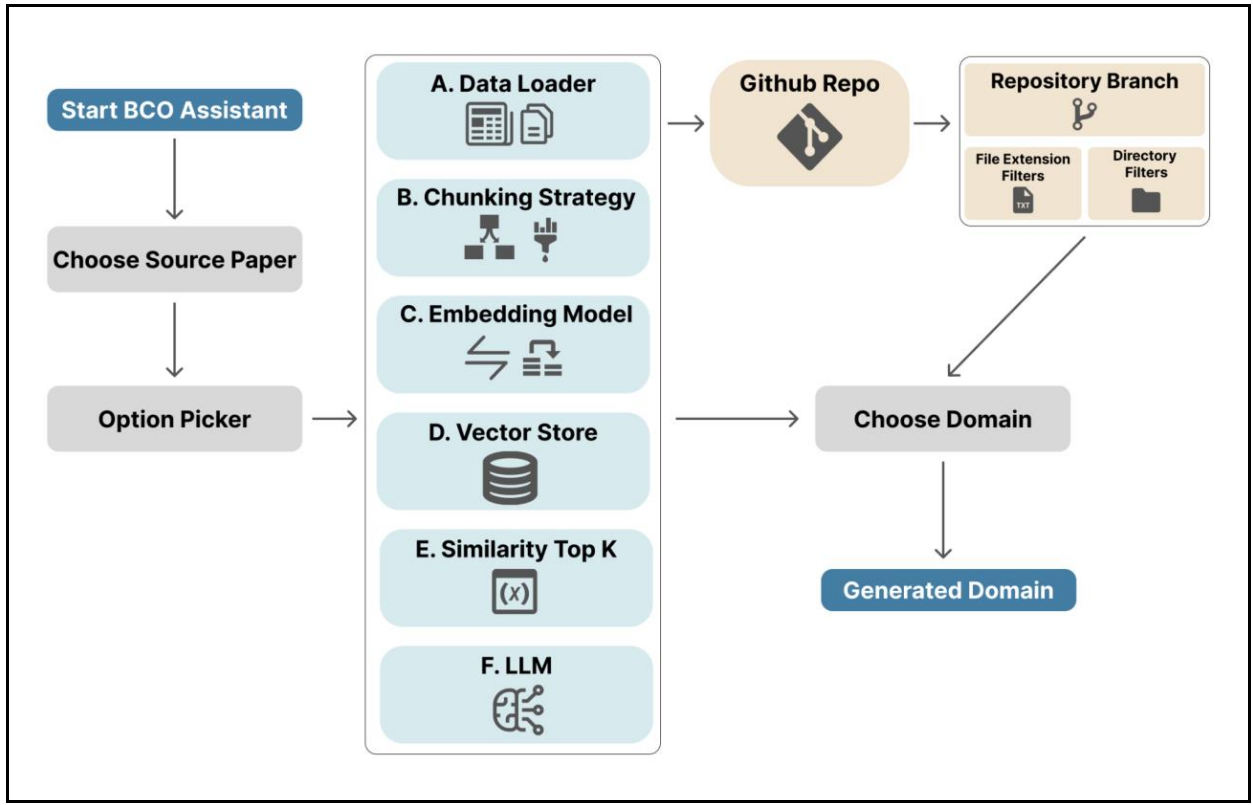